\documentclass{article} 
\usepackage{iclr2027_conference,times}

\usepackage{amsmath,amsfonts,bm}

\def\eqref#1{equation~\ref{#1}}

\def\1{\bm{1}}

\DeclareMathAlphabet{\mathsfit}{\encodingdefault}{\sfdefault}{m}{sl}
\SetMathAlphabet{\mathsfit}{bold}{\encodingdefault}{\sfdefault}{bx}{n}

\def\gV{{\mathcal{V}}}

\newcommand{\Ls}{\mathcal{L}}

\newcommand{\softmax}{\mathrm{softmax}}

\usepackage{hyperref}
\usepackage{url}
\usepackage{xspace}
\usepackage{enumitem}
\usepackage{graphicx}
\usepackage{algorithm}
\usepackage{algpseudocode}
\usepackage{wrapfig}
\usepackage{graphicx}   
\usepackage{booktabs}   
\usepackage{multirow}   
\usepackage{xcolor}     
\usepackage{relsize}    
\usepackage{xspace}     
\usepackage{array}      
\usepackage{tcolorbox}

\usepackage{etoolbox}
\makeatletter
\patchcmd{\@maketitle}
  {\lhead{Published as a conference paper at ICLR 2027}}
  {\lhead{Preprint.}}
  {}{}
\patchcmd{\@maketitle}
  {{\LARGE\sc \@title\par}}
  {{\fontsize{16}{19}\selectfont\centering \@title\par}}
  {}{}
\patchcmd{\@maketitle}
  {\begin{tabular}[t]{l}\bf\rule{\z@}{24pt}\@author\end{tabular}}
  {\centering\begin{tabular}[t]{c}\rule{\z@}{24pt}\@author\end{tabular}}
  {}{}
\makeatother

\newcommand{\sys}[0]{DPara\xspace}
\def\Snospace~{\S{}}
\newcommand{\parabf}[1]{\smallskip\noindent\textbf{#1}}
\newcommand{\tok}[1]{\fbox{\ensuremath{#1}}}
\newcommand{\tabsmall}{\relsize{-1}}

\newcommand{\takeaway}[1]{%
  \begin{tcolorbox}[
    colback=gray!6,
    colframe=black,
    boxrule=0.5pt,
    arc=0pt,
    outer arc=0pt,
    left=8pt,
    right=8pt,
    top=6pt,
    bottom=6pt,
    boxsep=0pt,
    before skip=6pt,
    after skip=6pt
  ]
  #1
  \end{tcolorbox}%
}

\title{When Parallel Drafter Meets Parallel Speculative Decoding}

\author{Fuliang Liu$^{1,2}$, \ Xue Li$^{2}$, \ Kun Qian$^{2}$, \ Zhibin Wang$^{1,*}${\expandafter\def\csname @makefnmark\endcsname{\hbox{}}\thanks{Corresponding author: Zhibin Wang (wzbwangzhibin@gmail.com.)}}\\[1mm]
Wanchun Dou$^{1}$, \ Wenyuan Yu$^{2}$, \ Chen Tian$^{1}$\\[2mm]
$^{1}$State Key Laboratory of Novel Software Technology, Nanjing University \quad $^{2}$Alibaba Group
}

\iclrfinalcopy 
\begin{document}

\maketitle

\vspace{-4mm}
\begin{abstract}
DSpark-style parallel drafters have made speculative decoding highly effective, yet their draft phase remains serialized on the critical path of every round. Parallel speculative decoding (PSD) overlaps drafting with verification, yet existing methods must guess the accepted prefix and bonus token in advance: a wrong guess reverts the whole batch to serial drafting.
We present \sys, a PSD framework that reuses effective parallel drafters yet guarantees backbone--verification overlap in every round, thereby eliminating this probabilistic fallback altogether. While the target verifies, \sys's diffusion backbone precomputes draft representations for \emph{every} acceptance boundary with the bonus left unspecified; a lightweight autoregressive head then combines the revealed verification outcome with the matching precomputed representation to emit the next round's draft tokens almost instantly---fully parallelizing the dominant backbone forward with verification and leaving only the negligible head cost serial.
Experiments on Qwen3-8B and Qwen3-14B across seven math, coding, and chat benchmarks show that \sys{} achieves average speedups of $3.21\times$ and $3.52\times$ over autoregressive decoding, surpassing the strongest serial and parallel speculative decoding baselines alike.
\end{abstract}

\begin{figure*}[h]
  \centering
  \includegraphics[width=\textwidth]{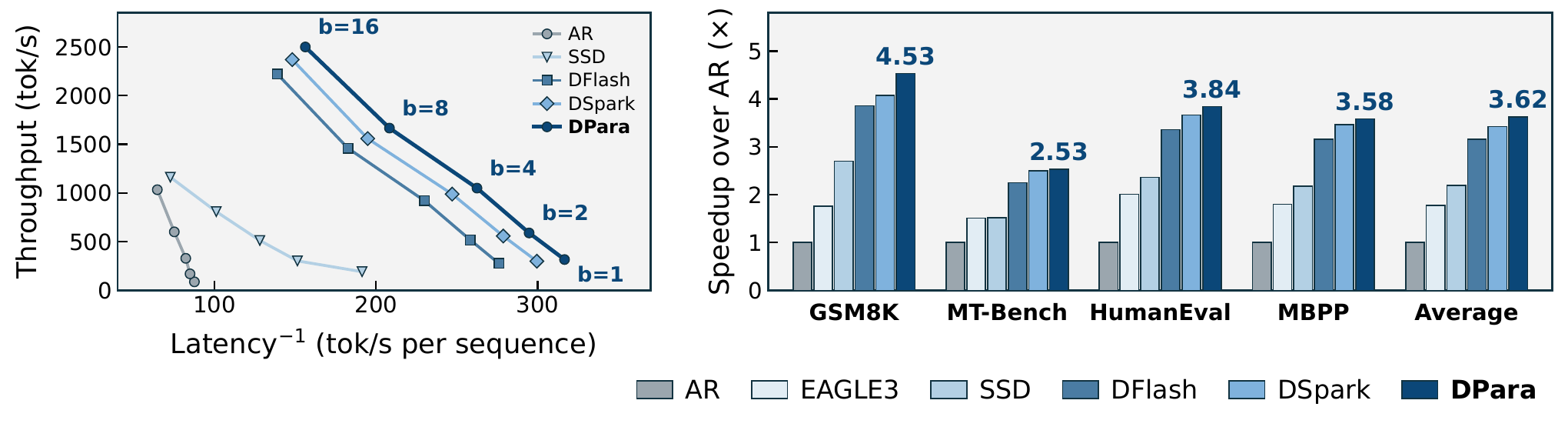}
  \vspace{-2em}
  \caption{Performance of \sys{} and other baselines on Qwen3-14B.
  \textbf{Left}: throughput--latency frontier, averaging over the four benchmarks shown on the right; each curve traces batch sizes $b{=}1,2,4,8,16$, with batch labels on \sys{}.
  \textbf{Right}: end-to-end speedup over AR at $b{=}1$; \sys{} reaches an arithmetic mean of $3.62\times$ across these four benchmarks, exceeding DSpark on each.}
  \label{fig:main-res}
\end{figure*}
\vspace{-4mm}
\section{Introduction}
\label{sec:introduction}

Speculative decoding (SD) accelerates large language model inference effectively
while preserving the target distribution~\citep{leviathan2023fast,chen2023accelerating}.
Its end-to-end efficiency is governed by two quantities: \emph{average acceptance length} and \emph{drafting latency} on the critical path.

\emph{Diffusion-inspired parallel drafters}, like DART~\citep{liu2026dart} and DFlash~\citep{chen2026dflash}, predict logits for multiple future positions
in a single forward pass and permit greater model capacity under the
same latency budget. 
Parallel logit prediction alone, however, does not capture dependencies among the sampled draft tokens.
Subsequent designs such as DSpark~\citep{cheng2026dspark} and
Domino~\citep{huang2026domino} address this limitation by coupling the parallel
diffusion backbone with a lightweight autoregressive head, which restores sequence consistency. Yet even under these advanced drafters, the conventional sequential
draft--verify schedule keeps drafting on the critical path, where it accounts
for $24\%$--$32\%$ of end-to-end latency (Appendix~\ref{app:dflash_time_breakdown}).

\begin{figure*}[t]
  \centering
  \includegraphics[width=0.85\textwidth]{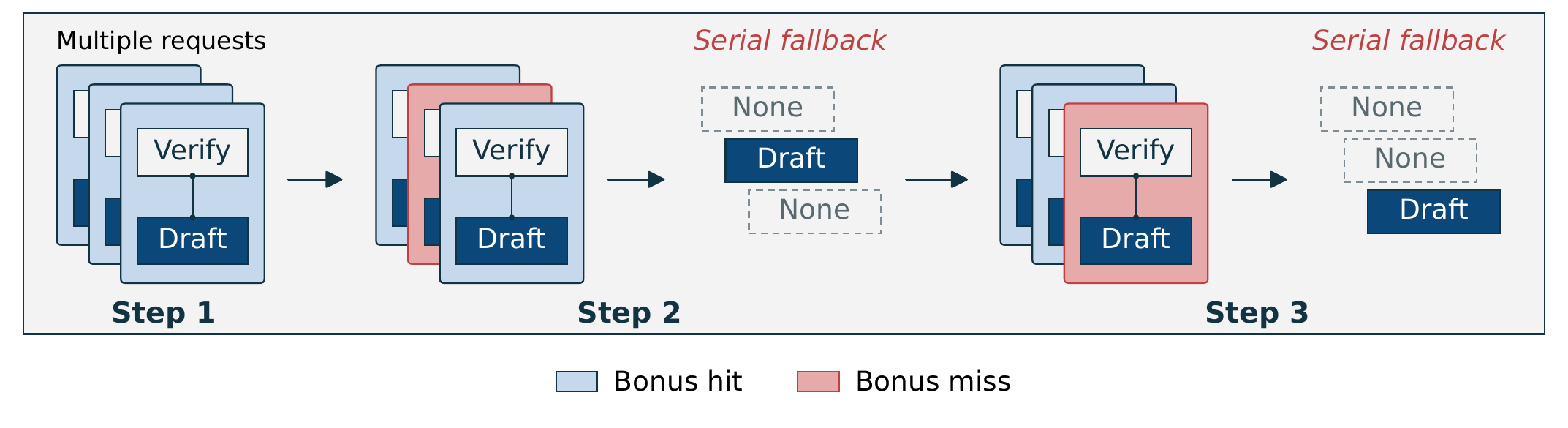}
  \vspace{-1em}
  \caption{SSD drafts and verifies a batch of requests together. When any request's predicted bonus token mismatches (red), the whole batch loses its precomputed continuation and reverts to serial drafting. As the batch grows, the chance of at least one mismatch rises, so larger batches fall back more often and forfeit the parallel draft--verify gain.}
  \label{fig:larger-bs}
  \vspace{-4mm}
\end{figure*}

\emph{Parallel speculative decoding} (PSD) attacks this serialization directly
by overlapping draft generation with target
verification~\citep{pearl,amusd,specbranch,ssd}. On behalf of these methods, SSD~\citep{ssd} extends speculation to verification itself: while the
target verifies the current step, the drafter predicts the accepted
prefix and \emph{bonus token} and generates subsequent draft tokens in advance.
A matching outcome makes the next step's draft tokens immediately available,
eliminating drafting latency from the critical path. A mismatch, however, reverts SSD to serial speculative decoding (Figure~\ref{fig:larger-bs}), and because the probability of at least one mismatch grows with batch size, the entire batch is forced back to serial execution ever more frequently---at batch size $16$, up to $96\%$ of steps fall back (Appendix~\ref{app:ssd-fallback}). Outcome-dependent overlap therefore exchanges a best-case gain for a regression that intensifies with serving load. This tension raises a natural question:
\begin{center}
\vspace{-2mm}
\emph{Can we \textbf{completely eliminate probable serial fallback} from parallel speculative decoding?}
\vspace{-2mm}
\end{center}

We answer this question affirmatively by severing the probabilistic dependence on
bonus-token prediction: the key is to separate the computation that must await
the verification outcome from the computation that can proceed independently of it.
The DSpark-style architecture affords precisely this separation---the expensive diffusion backbone computes draft representations
while verification is still in flight, and the lightweight autoregressive head
incorporates the actual bonus token afterward. Deferring bonus-token
conditioning to the head yields \textbf{backbone--verification overlap in
every round, without any prediction-induced fallback}.

We present \sys, a PSD framework that makes this overlap independent of
the verification outcome. \sys prepares diffusion backbone prediction outputs for \emph{every}
possible accepted prefix boundary and leaves the bonus token unspecified until
verification completes. Because this guarantee holds for
each request individually, enlarging the batch cannot reintroduce
the prediction-induced serial fallback.
Covering all acceptance boundaries in a single pass, however, requires a
backbone that predicts from multiple feature-less anchors---a regime the
standard DFlash model does not satisfy. We close this gap by finetuning:
M-DFlash initializes from the published DSpark checkpoint and adapts
\emph{only the diffusion backbone} to the multiple-anchor, feature-less
regime, introducing no new trainable parameters and keeping the AR head frozen
as published.
Provisioned to complete within the verification window, the diffusion backbone---the dominant component of drafting latency---is fully hidden, leaving only the lightweight autoregressive head on the critical path. \sys thereby preserves the effectiveness of parallel drafters while excising backbone execution entirely from the serial draft--verify schedule.

Figure~\ref{fig:main-res} previews the empirical gains. Across math,
coding, and chat benchmarks on Qwen3-8B and Qwen3-14B, \sys{} exceeds other baselines on every model--dataset pair, averaging $3.21\times$ and $3.52\times$ speedup over autoregressive decoding (\S\ref{sec:eval:main}).
Against DSpark, the strongest serial drafter, this corresponds to average
improvements of $11.1\%$ and $5.1\%$; against the parallel baselines PEARL
and SSD on Qwen3-14B, the average speedup rises by $31\%$ and $64\%$.
The advantage does not erode under batching: \sys{} stays ahead of DSpark at
every batch size from $1$ to $16$ and retains over $2\times$ AR throughput at
batch $16$, while outcome-dependent methods such as SSD suffer a markedly steeper throughput drop-off as batches grow (Figure~\ref{fig:main-res}).
The gains likewise survive a leaner drafting budget: \sys{}-A10 retains $98\%$ of
\sys{}'s speedup while cutting the drafting-device cost by ${\sim}90\%$,
whereas \sys{}-single retains $90\%$--$95\%$ at small batches
(\S\ref{sec:eval:futher}).
Ablations further confirm that finetuning only the diffusion backbone raises the average acceptance length by $30.7\%$.

This paper makes the following contributions:
\begin{itemize}[leftmargin=1.2em,itemsep=1pt,topsep=2pt]
  \item The \textbf{first} parallel speculative decoding framework
    powered by DSpark-style parallel drafters, establishing
    \textbf{state-of-the-art} inference performance across both serial and
    parallel baselines.
  \item \sys, a PSD framework that covers all acceptance
    boundaries \textbf{without any serial backbone fallback} and hides the backbone
    latency entirely within the verification window, enabled by M-DFlash, a
    multiple-anchor backbone finetuned from a published DSpark checkpoint.
  \item An extensive evaluation on Qwen3-8B and Qwen3-14B: \sys{}
    exceeds the best serial and parallel baselines on every model--dataset
    pair (up to $4.53\times$, averaging $3.21\times$/$3.52\times$ over AR) and sustains its advantage from batch $1$ to $16$; further explorations of
    heterogeneous drafting and single-GPU colocation
    retain most of the speedup under a reduced
    drafting budget.
\end{itemize}

\vspace{-2mm}
\section{Background and Related Work}
\label{sec:background}

\begin{figure*}[t]
  \centering
  \includegraphics[width=0.75\textwidth]{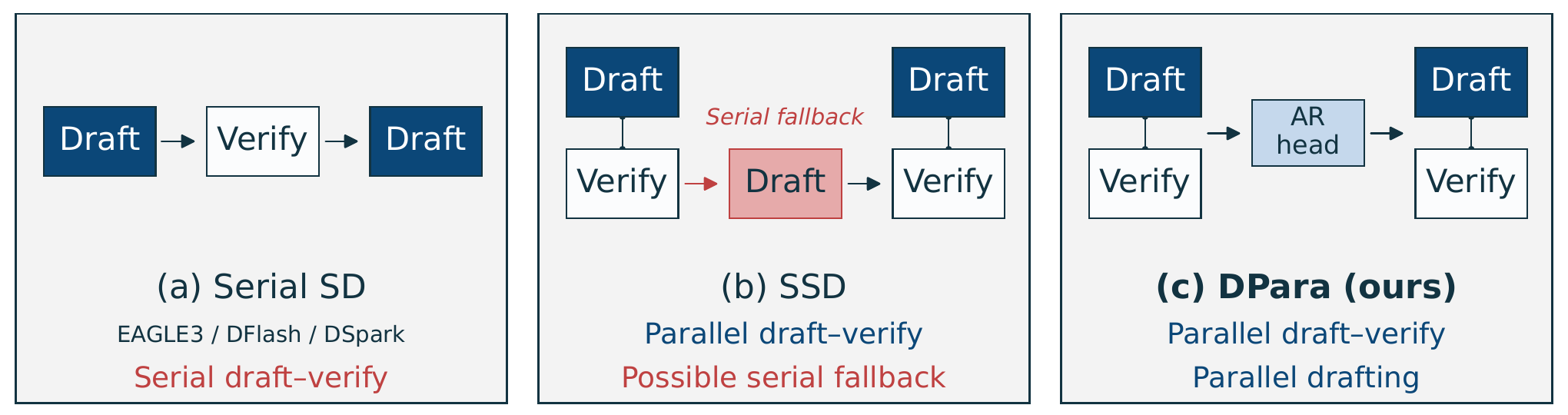}
  \caption{Scheduling of drafting and verification across speculative decoding paradigms.
  \textbf{(a) Serial SD}: autoregressive (EAGLE3) or parallel drafters (DFlash/DSpark) run draft--verify in sequence, leaving drafting on the critical path.
  \textbf{(b) SSD}: overlaps drafting with verification, but a mispredicted bonus token forces serial fallback.
  \textbf{(c) \sys{} (ours)}: the backbone overlaps verification for every acceptance outcome while drafting stays parallel.}
  \label{fig:bg}
  \vspace{-4mm}
\end{figure*}

\subsection{Speculative Decoding}
\label{sec:bg:llm}
\vspace{-2mm}

Speculative decoding~\citep{leviathan2023fast,chen2023accelerating}
alleviates memory-bound decoding by amortizing the weight reads of a
large target model $\mathcal{M}_t$ across multiple tokens. Given a prefix
$C$, a lightweight drafter $\mathcal{M}_d$ proposes a block
$\tilde{y}_{1:K}\sim q(\cdot\mid C)$, which $\mathcal{M}_t$ scores in one
forward pass. Distribution-preserving rejection sampling retains an
accepted prefix and appends a \emph{bonus token} $b$: a correction sampled
at the first rejection, or an additional target sample after full
acceptance.
\vspace{-2mm}

\subsection{Parallel Drafter}
\label{sec:bg:pd}
\vspace{-2mm}

To overcome the sequential drafting overhead of autoregressive drafters
such as EAGLE3~\citep{li2025eagle3}, DART~\citep{liu2026dart} and
DFlash~\citep{chen2026dflash} draw inspiration from diffusion language models (dLLMs) and predict a block of future positions in one forward pass, substantially reducing draft-generation cost. Independent prediction at different positions, however, can produce
incoherent draft sequences and declining acceptance length ($\tau$). DART applies \emph{N-gram} corrections to its parallel logits,
whereas Domino~\citep{huang2026domino} and DSpark~\citep{cheng2026dspark}
learn a lightweight autoregressive head that conditions each position on
the preceding sampled tokens, which preserves the efficient parallel
backbone while restoring local causal dependence, leading to longer $\tau$. Nevertheless, the drafter and verifier remain sequentially coupled across
decoding steps: \textbf{the drafter must finish before verification can begin,
leaving drafting on the critical path and making it a nontrivial component
of end-to-end latency}.
\vspace{-2mm}

\subsection{Parallel speculative decoding}
\label{sec:bg:psd}
\vspace{-2mm}
Figure~\ref{fig:bg} contrasts the execution schedules of serial SD,
SSD, and \sys{}.
PSD overlaps next-step drafting with current-step verification to hide
the drafting interval that otherwise delays the next verification.
The drafter must therefore generate tentative continuations before the
accepted prefix and \emph{bonus token} are known.
PEARL~\citep{pearl} and AMUSD~\citep{amusd} generate continuations during verification under an all-accepted assumption, which provides only limited opportunities for parallelism.
SpecBranch~\citep{specbranch} and SSD~\citep{ssd} predict likely rejection positions and
bonus tokens to reduce rollback.
In all these methods, an unanticipated verification outcome can invalidate
the precomputed continuation and fall back to \emph{serial speculation},
\textbf{so none of them guarantees draft--verify parallelism without fallback in every round}.

\vspace{-4mm}
\section{\sys Design}
\vspace{-3mm}
In this section, we will go through the design of \sys to guarantee complete backbone--verification overlap in every PSD step.
\vspace{-3mm}
\begin{figure*}[t]
  \centering
  \includegraphics[width=\textwidth]{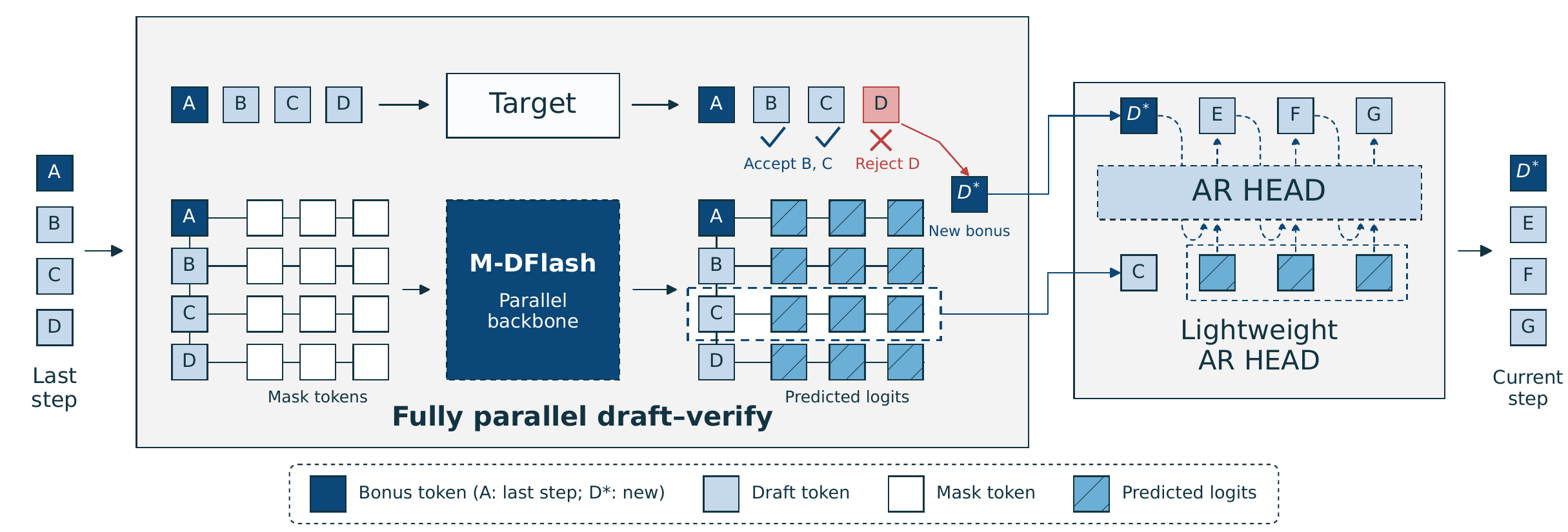}
  \vspace{-4mm}
  \caption{One decoding step of \sys. With \tok{A} as the bonus token from last step,
  the target verifies \tok{ABCD} while the parallel backbone concurrently precomputes $d$ future logits for every acceptance boundary. Accepting \tok{BC} and rejecting \tok{D} yields
  bonus token \tok{D^*}. The lightweight AR head combines \tok{D^*} with the
  logits following \tok{C} to generate next step's draft tokens \tok{EFG} quickly.}
  \vspace{-5mm}
  \label{fig:design-detail}
\end{figure*}


\subsection{Backbone requirement of \sys}
\sys places a requirement on the \emph{diffusion backbone} that the standard DFlash
model does not satisfy. Under the conventional serial schedule, drafting runs
\emph{after} verification: the target has already finished verification, so a
fresh target feature exists at every position before bonus token---\emph{the bonus itself carries no feature of its own}. DFlash~\citep{chen2026dflash} anchors on this \emph{single} token, appended
to the fully featured prefix, to predict the logits of the next $d$ positions in
one backbone forward. \sys instead launches the backbone \emph{while} the target is
still verifying the current block (Figure~\ref{fig:design-detail}). Write the
draft tokens from last step as $A,B,C,D$, where $A$ is the previous bonus and
$B,C,D$ are the drafts being scored, and let $r\in\{0,\dots,d\}$ be the
number of accepted drafts and $b$ the new bonus. At launch time the backbone
knows neither $r$ nor $b$; all it holds are the tokens carried over from the
previous round, of which the accepted prefix $A,t_1\dots,t_r$ forms
$k=r+1\ (\ge 1)$ anchor tokens that have \emph{not} yet passed through the
target and therefore carry \emph{no} target features. This requires: 
\takeaway{The \sys's backbone must predict the next $d$ positions from \textbf{multiple, feature-less anchors}, and must do so for every possible $r$ at once, since the true acceptance length is not revealed until verification finishes.}

\begin{wrapfigure}{r}{0.44\textwidth}
\centering
\vspace{1em}
\includegraphics[width=0.38\textwidth]{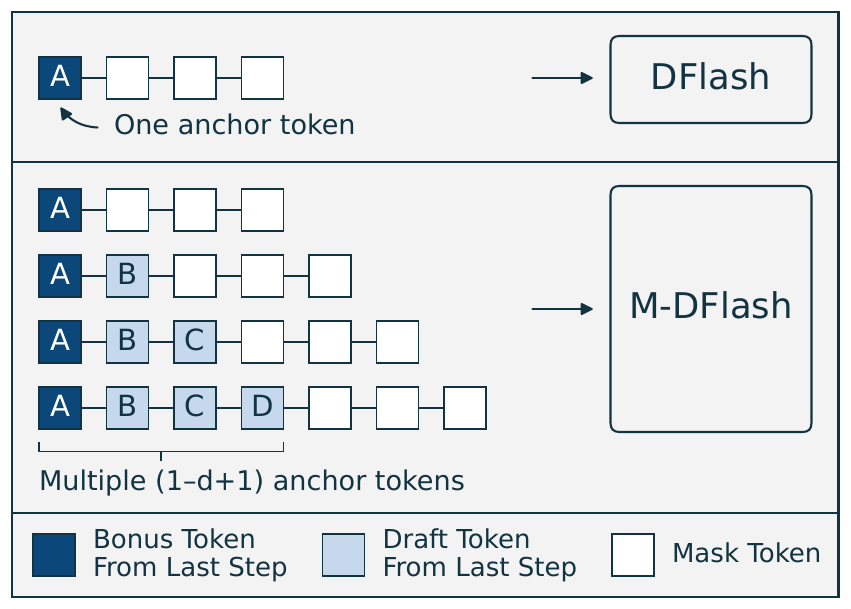}
\caption{Input regimes of DFlash and M-DFlash. \textbf{Top}: DFlash anchors on a single forwarded token appended to a fully featured prefix. \textbf{Bottom}: M-DFlash predicts the next $d$ logits from multiple ($1{\sim}d{+}1$) feature-less anchor tokens for every acceptance boundary in one forward pass---the regime required to launch the backbone during verification.}
\label{fig:anchors-vs-multi}
\end{wrapfigure}

This requirement, however, does not change \emph{what} the backbone computes; it
\emph{generalizes} it. DFlash already predicts future logits from \emph{a
single anchor} given the fully featured prefix; \sys only asks it to extend this \emph{from a single anchor to multiple anchors}. Because multiple-anchor prediction is a strict extension of DFlash's single-anchor semantics rather than a different function, we can obtain the backbone \sys needs by finetuning DFlash (Figure~\ref{fig:anchors-vs-multi}). We call this multiple-anchor diffusion backbone \textbf{M-DFlash} (Multiple-anchor-token DFlash).

\vspace{-4mm}
\subsection{Finetuning DFlash to M-DFlash}
\label{sec:design:train}

M-DFlash introduces \emph{no new trainable network parameters}. We initialize it from
the official DSpark checkpoint~\citep{cheng2026dspark}---a DFlash-family drafter
that couples the block-diffusion backbone with a lightweight Markov
autoregressive (AR) head---and adapt \emph{only the backbone part} to the multiple-anchor,
feature-less regime above. The AR head is precisely the component that lets
\sys defer bonus conditioning to the post-verification stage, so we keep it
frozen as published.

\begin{algorithm}[t]
\caption{One steady-state round of \sys (draft length $d$).}
\label{alg:dssd}
\begin{algorithmic}[1]
\Require feature cache $C$; feature-less draft block $t_{0:d}$ ($t_0$ = previous bonus)
\Ensure accepted tokens, bonus $b$; updated $C$ and next block
\Statex \textit{// launched together, run concurrently}
\State \textbf{Precompute} (drafter): M-DFlash on spine $t_{0:d}$, cache $C$
\Statex \quad $\rightarrow$ base logits $\ell^{(r)}_{1:d}$ for all $r\in\{0{:}d\}$
\State \textbf{Verify} (target): $\mathcal{M}_t$ forwards $t_{0:d}$
\Statex \quad $\rightarrow$ accepted length $r$, bonus $b$, features of $t_{0:r}$
\Statex \textit{// barrier: $(r,b)$ now known}
\State select precomputed branch $r$ \Comment{enumerated $\Rightarrow$ no fallback}
\State $y_0 \gets b$
\For{$i=1,\dots,d$} \Comment{AR head: lightweight cost}
  \State $z_i \gets \ell^{(r)}_i + W E(y_{i-1})$;\quad $y_i\sim\softmax(z_i)$
\EndFor
\State $C \gets [\,C,C^{new}_{0:r}\,]$ \Comment{drop rejected $t_{r+1:d}$}
\State $t_{0:d} \gets (b,\,y_{1:d})$ \Comment{$b$ = next feature-less anchor}
\State \Return accepted $t_{1:r}$, bonus $b$ \Comment{stop at EOS / length budget}
\end{algorithmic}
\end{algorithm}

\parabf{Trainable and frozen parameters.} We train the backbone blocks
(attention, MLP, and their normalizations), the projection that fuses the
concatenated multi-layer target features, and the input and output
normalizations. We freeze the entire target $\mathcal{M}_t$, which only supplies
online features and supervision, together with the drafter's token embedding,
its full-vocabulary LM projection, and both the AR head and the confidence head.
Only the backbone faces a changed input regime---\emph{multiple feature-less anchors
instead of a single one}---so it is the only part that must adapt.

\begin{wrapfigure}{r}{0.53\textwidth}
\centering
\vspace{-3em}
\includegraphics[width=0.52\textwidth]{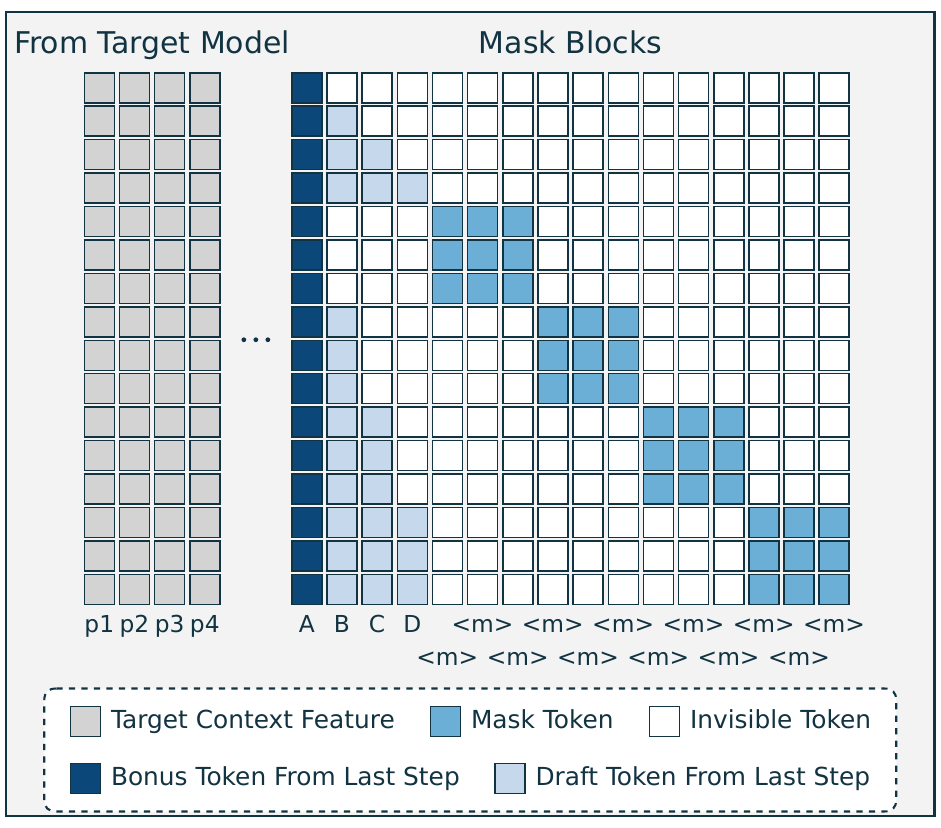}
\caption{M-DFlash attention mask for $d=3$.
Colored cells indicate visible
connections and white indicate masked connections.
All queries access cached target features (gray).
The previous bonus $A$ (dark blue) and drafts $B,C,D$ (light blue)
form a shared causal spine.
Four bidirectional three-mask branches (medium blue) read the anchor prefixes
\tok{A}, \tok{AB}, \tok{ABC}, and \tok{ABCD}, respectively, without attending to one another.}
\label{fig:mask-blocks}
\end{wrapfigure}

\parabf{Shared causal spine with isolated bidirectional branches.}
M-DFlash covers all $d+1$ acceptance boundaries in one backbone forward
using the attention pattern in Figure~\ref{fig:mask-blocks}.
In the illustrated example ($d=3$), the previous bonus $A$ and draft
tokens $B,C,D$ form a shared, feature-less causal spine.
Each anchor attends to the cached target features, itself, and earlier
anchors, but never to later anchors or mask tokens.
The four groups of three masks correspond to the possible accepted
prefixes \tok{A}, \tok{AB}, \tok{ABC}, and \tok{ABCD}.
Within each group, the masks attend bidirectionally to one another,
read the cached target features, and access only the corresponding
anchor prefix; all other branches remain invisible.
For example, the branch following $C$ reads $A,B,C$ but not $D$,
so its predictions remain valid if verification accepts $B,C$ and
rejects $D$.
The triangular light blue region and diagonal medium blue blocks therefore encode
two complementary constraints: causal sharing along the spine and
bidirectional attention within isolated branches.
Unlike an ordinary triangular causal mask, this tree-structured mask
computes the shared anchors once while producing $d$ future logits
for every possible acceptance boundary.

\parabf{Loss.} For a selected boundary, let $\ell_i$ be the base logits at mask
position $i$ and let the frozen AR head add a correction from the previously
sampled token,
\[
z_i = \ell_i + W E(y_{i-1}), \qquad y_0 = b, \quad i = 1,\dots,d,
\]
where $E$ is the head's independent low-rank token embedding and $W$ projects it
to the vocabulary; the head reads only the preceding token. Writing $q_i=\softmax(z_i)$ for the student and $p_i$ for the
target distribution at the same position, we minimize a position-weighted
mixture of cross-entropy and full-vocabulary probability distance~\citep{cheng2026dspark},
\[
\Ls = \frac{\sum_{i} m_i\, w_i \left[\, 0.1\,\big(-\log q_i(y_i)\big) + 0.9 \sum_{v\in\gV} \lvert q_i(v) - p_i(v) \rvert \right]}{\sum_i m_i\, w_i}, \qquad w_i = e^{-(i-1)/\gamma},
\]
with decay $\gamma=4$ and $m_i$ a validity mask that keeps only positions whose
supervision falls inside the assistant span. The distribution-matching term
aligns the drafter's proposal with the target's acceptance behavior over the
whole vocabulary; the confidence loss is disabled, so the fixed budget $d$ is
never dynamically truncated.

\subsection{\sys Algorithm}
\label{sec:design:alg}

Algorithm~\ref{alg:dssd} details one steady-state round. \sys maintains a
target-feature cache $C$ over the accepted history and a feature-less draft
block $t_{0:d}$ carried over from the previous round, where $t_0$ is the previous
bonus and $t_{1:d}$ are the drafts now being scored. The M-DFlash backbone
runs on the shared spine $t_{0:d}$ concurrently with the target verifying that
same token block, and emits base logits for \emph{all} $d+1$ acceptance boundaries. This is exactly where \sys departs from
SSD~\citep{ssd}, which predicts the accepted prefix and bonus token to
pre-generate the next draft and reverts to serial speculation whenever that
prediction misses. \sys instead predicts future logits, and it never has to
\emph{guess the verification outcome}: it commits to no particular accepted prefix or particular bonus token,
so whatever $(r,b)$ the verifier returns, a finished branch is already waiting.

\parabf{Only the head stays on the critical path.} When verification returns
$(r,b)$, \sys selects branch $r$ and runs \emph{only} the lightweight AR head to turn
its corresponding predicted logits into the next draft $y_{1:d}$; since the backbone already overlapped with verification, this low-rank correction is the sole additional cost
exposed between two rounds. The verified features of $t_{0:r}$ are appended to
$C$ and the rejected suffix dropped, while the bonus $b$---sampled at the last
accepted position and therefore carrying no target feature \emph{yet}---becomes
the feature-less $t_0$ of the next round, to be forwarded until next verification. Approximating a round as
\[
T_{\text{step}}\approx\max\!\big(T_{\text{verify}},T_{\text{precompute}}\big)+T_{\text{Head}},
\]
the diffusion backbone term is hidden whenever it fits inside the verification window.

\section{Experiments}
\label{sec:evaluation}

\subsection{Experiments Setup}
\label{sec:eval:setup}
\paragraph{Target models.} 
We train M-DFlash drafters for two models of the Qwen3 family (Qwen3-8B and Qwen3-14B~\citep{qwen3}); the detailed training recipe is provided in Appendix~\ref{app:training-recipe}.

\paragraph{Hardware.} All training is conducted on a server equipped with NVIDIA GB200 GPUs (4 GPUs, 268\,GB). The main results are obtained on an NVIDIA H800 node (8 GPUs, 80\,GB each). We also conduct additional inference experiments on NVIDIA GB200 GPUs. For PSD methods, including PEARL, SSD, and \sys{}, we use one additional GPU dedicated to drafting. We further explore an NVIDIA A10 ($1\times24$GB) as a resource-constrained drafting device (\S\ref{sec:eval:futher}).

\paragraph{Benchmarks.}
We evaluate \sys on seven datasets across three workload categories:
\textbf{Math} for mathematical reasoning, including GSM8K~\citep{gsm8k}
and MATH-500~\citep{Math500};
\textbf{Chat} for conversation and instruction following, including
MT-Bench~\citep{mt_bench} and Alpaca~\citep{alpaca};
and \textbf{Coding} for code generation, including
HumanEval~\citep{humaneval}, MBPP~\citep{MBPP}, and
CodeAlpaca~\citep{CodeAlpaca}.

\subsection{Main Result}
\label{sec:eval:main}

\begin{table*}[t]
    \vspace{-5mm}
    \caption{\sys against parallel (PSD) and serial (SD) speculative decoding baselines on Qwen3-8B and Qwen3-14B. Within each model block, the upper group of rows lists PSD baselines, the middle group serial SD baselines, and the last row \sys. We report decoding speedup over the AR baseline and the average acceptance length $\tau$, with batch size $1$, greedy decoding, up to $512$ new tokens, draft length $7$. SSD-E3 takes corresponding EAGLE3 weight of target as draft model. SSD and PEARL in Qwen3-14B take Qwen3-0.6B as draft model.}
    \vspace{2mm}
    \label{tab:main}
    \resizebox{\linewidth}{!}{
        \normalsize
        \centering
        \setlength{\tabcolsep}{2pt}
        \begin{tabular}{c l @{\hspace{0.6em}} cc cc @{\hspace{0.8em}} cc cc cc @{\hspace{0.8em}} cc cc @{\hspace{0.8em}} cc}
            \toprule
            \multirow{2}{*}{Model} & \multirow{2}{*}{Method} & \multicolumn{4}{c@{\hspace{0.8em}}}{\sc{Math}} & \multicolumn{6}{c@{\hspace{0.8em}}}{\sc{Coding}} & \multicolumn{4}{c@{\hspace{0.8em}}}{\sc{Chat}} & \multicolumn{2}{c}{} \\
            \cmidrule(lr){3-6} \cmidrule(lr){7-12} \cmidrule(lr){13-16}
            & & \multicolumn{2}{c}{GSM8K} & \multicolumn{2}{c@{\hspace{0.8em}}}{MATH-500} & \multicolumn{2}{c}{HumanEval} & \multicolumn{2}{c}{MBPP} & \multicolumn{2}{c@{\hspace{0.8em}}}{CodeAlpaca} & \multicolumn{2}{c}{Alpaca} & \multicolumn{2}{c@{\hspace{0.8em}}}{MT-Bench} & \multicolumn{2}{c}{\sc{Avg.}} \\
            & & \tabsmall Speedup & \tabsmall $\tau$ & \tabsmall Speedup & \tabsmall $\tau$ & \tabsmall Speedup & \tabsmall $\tau$ & \tabsmall Speedup & \tabsmall $\tau$ & \tabsmall Speedup & \tabsmall $\tau$ & \tabsmall Speedup & \tabsmall $\tau$ & \tabsmall Speedup & \tabsmall $\tau$ & \tabsmall Speedup & \tabsmall $\tau$ \\
            \midrule
            \multirow{5}{*}{Q3-8B}
            & SSD-E3
            & 1.51$\times$ & 2.39 & 1.54$\times$ & 2.42 & 1.72$\times$ & 2.75 & 1.58$\times$ & 2.49 & 1.67$\times$ & 2.62 & 1.29$\times$ & 2.17 & 1.34$\times$ & 2.26 & 1.52$\times$ & 2.44 \\
            \cmidrule(lr){2-18}
            & EAGLE3
            & 1.44$\times$ & 2.83 & 1.46$\times$ & 2.84 & 1.67$\times$ & 3.26 & 1.51$\times$ & 2.98 & 1.58$\times$ & 3.18 & 1.18$\times$ & 2.45 & 1.27$\times$ & 2.57 & 1.44$\times$ & 2.87 \\
            & DFlash
            & 3.38$\times$ & 6.08 & 3.28$\times$ & 5.88 & 2.93$\times$ & 5.25 & 2.80$\times$ & 4.95 & 2.77$\times$ & 4.95 & 1.81$\times$ & 3.28 & 1.94$\times$ & 3.56 & 2.70$\times$ & 4.85 \\
            & DSpark
            & 3.48$\times$ & 6.51 & 3.46$\times$ & 6.39 & 3.15$\times$ & 5.82 & 3.00$\times$ & 5.51 & 3.01$\times$ & 5.55 & 2.00$\times$ & 3.72 & 2.13$\times$ & 4.01 & 2.89$\times$ & 5.36 \\
            \cmidrule(lr){2-18}
            & \textbf{\sys}
            & \textbf{4.17$\times$} & 5.82 & \textbf{3.95$\times$} & 5.41 & \textbf{3.50$\times$} & 4.78 & \textbf{3.27$\times$} & 4.48 & \textbf{3.25$\times$} & 4.47 & \textbf{2.10$\times$} & 2.92 & \textbf{2.25$\times$} & 3.12 & \textbf{3.21$\times$} & 4.43 \\
            \midrule
            \multirow{8}{*}{Q3-14B}
            & SSD
            & 2.70$\times$ & 5.47 & 2.80$\times$ & 5.63 & 2.36$\times$ & 4.87 & 2.18$\times$ & 4.47 & 2.05$\times$ & 4.39 & 1.39$\times$ & 3.17 & 1.51$\times$ & 3.47 & 2.14$\times$ & 4.50 \\
            & SSD-E3
            & 1.76$\times$ & 2.48 & 1.80$\times$ & 2.48 & 2.02$\times$ & 2.80 & 1.82$\times$ & 2.54 & 1.91$\times$ & 2.68 & 1.48$\times$ & 2.17 & 1.57$\times$ & 2.28 & 1.77$\times$ & 2.49 \\
            & PEARL
            & 3.35$\times$ & 3.70 & 3.58$\times$ & 3.85 & 2.95$\times$ & 3.18 & 2.47$\times$ & 2.81 & 2.58$\times$ & 2.81 & 1.84$\times$ & 1.98 & 2.03$\times$ & 2.20 & 2.68$\times$ & 2.93 \\
            \cmidrule(lr){2-18}
            & SD
            & 2.21$\times$ & 5.47 & 2.28$\times$ & 5.65 & 1.99$\times$ & 4.86 & 1.85$\times$ & 4.46 & 1.76$\times$ & 4.39 & 1.31$\times$ & 3.17 & 1.42$\times$ & 3.47 & 1.83$\times$ & 4.50 \\
            & EAGLE3
            & 1.76$\times$ & 2.94 & 1.74$\times$ & 2.93 & 2.01$\times$ & 3.35 & 1.80$\times$ & 3.04 & 1.86$\times$ & 3.24 & 1.42$\times$ & 2.42 & 1.51$\times$ & 2.57 & 1.73$\times$ & 2.93 \\
            & DFlash
            & 3.86$\times$ & 6.03 & 3.76$\times$ & 5.92 & 3.36$\times$ & 5.20 & 3.16$\times$ & 4.89 & 3.13$\times$ & 4.97 & 2.07$\times$ & 3.24 & 2.25$\times$ & 3.55 & 3.08$\times$ & 4.83 \\
            & DSpark
            & 4.07$\times$ & 6.53 & 4.01$\times$ & 6.39 & 3.66$\times$ & 5.79 & 3.46$\times$ & 5.48 & 3.45$\times$ & 5.60 & \textbf{2.32$\times$} & 3.71 & 2.50$\times$ & 4.02 & 3.35$\times$ & 5.36 \\
            \cmidrule(lr){2-18}
            & \textbf{\sys}
            & \textbf{4.53$\times$} & 5.73 & \textbf{4.32$\times$} & 5.37 & \textbf{3.84$\times$} & 4.76 & \textbf{3.58$\times$} & 4.45 & \textbf{3.49$\times$} & 4.46 & \textbf{2.32$\times$} & 2.91 & \textbf{2.53$\times$} & 3.16 & \textbf{3.52$\times$} & 4.41 \\
            \bottomrule
        \end{tabular}
    }
\end{table*}

\paragraph{Gains over both serial and parallel speculation.}
\sys{} achieves average speedups over autoregressive decoding of $3.21\times$ on Qwen3-8B and
$3.52\times$ on Qwen3-14B across seven benchmarks at batch size $1$
(Table~\ref{tab:main}).
It attains the highest reported speedup on every model--dataset pair,
with a tie with DSpark on Qwen3-14B Alpaca.
The advantage spans math, coding, and chat, and extends beyond methods
that serialize drafting and verification: on Qwen3-14B, \sys{} improves
the average speedup over PEARL and SSD by $31\%$ and $64\%$, respectively.
These results show that effective backbone--verification overlap
translates into lower end-to-end decoding latency.
Across the evaluated models and workloads, \sys{} achieves
state-of-the-art speedups across both serial and parallel
speculative decoding methods.

\paragraph{Balancing draft acceptance and critical-path cost.}
\sys{} is designed around a joint optimization of acceptance length and
drafting cost for end-to-end latency: it preserves comparatively high acceptance through a
high-capacity diffusion backbone while hiding the backbone forward pass
entirely within verification, leaving only the lightweight AR head on
the drafting critical path. This balance yields the best overall
end-to-end decoding latency among the evaluated methods.
Specifically, \sys{} retains average acceptance lengths of $4.43$ and
$4.41$ on Qwen3-8B and Qwen3-14B, respectively, compared with $5.36$ for
DSpark, the strongest serial SD baseline (Table~\ref{tab:main}).
This acceptance gap reflects the information constraints imposed by PSD:
when the backbone starts, neither the accepted length nor the new bonus
is known, and fresh target features for the draft tokens carried over
from the previous round are unavailable.
Unlike serial drafting, which conditions on a resolved acceptance
boundary and a fully featured prefix, \sys{} must precompute all
boundaries from feature-less anchors and defer bonus conditioning to
the AR head. The resulting information deficit constrains the
acceptance achievable by the parallel backbone relative to serial
drafting.
Nevertheless, the end-to-end results show that removing backbone
execution from the critical path more than compensates for this
acceptance gap: \sys{} improves average speedup over DSpark by $11.1\%$
on Qwen3-8B and $5.1\%$ on Qwen3-14B.
The net benefit is lower latency per generated token, achieved by
retaining strong draft acceptance without paying for a serial
backbone forward pass.

\begin{wraptable}{r}{0.5\textwidth}
    \centering
    \normalsize
    \setlength{\tabcolsep}{1.2pt}
    \setlength{\intextsep}{4pt}
    \renewcommand{\arraystretch}{0.90}
    \vspace{-9mm}
    \caption{Batched inference on Qwen3-8B/14B (Q3-8B/14B). Each method reports throughput (tok/s) and speedup over AR.}
    \vspace{1mm}
    \resizebox{\linewidth}{!}{
        \begin{tabular}{@{}c@{\hspace{0.3em}}l@{\hspace{0.3em}}l@{\hspace{0.3em}}ccccc@{}}
            \toprule
            \multirow{2}{*}{Model} & \multirow{2}{*}{Task} & \multirow{2}{*}{Method} & \multicolumn{5}{c@{}}{Batch size} \\
            \cmidrule{4-8}
            & & & 1 & 2 & 4 & 8 & 16 \\
            \midrule
            \multirow{10}{*}[-1.2ex]{Q3-8B}
            & \multirow{5}{*}[-0.5ex]{GSM8K} & AR & 141 & 277 & 529 & 937 & 1520 \\
            \cmidrule{3-8}
            & & \multirow{2}{*}{DSpark} & 491 & 901 & 1507 & 2426 & 3627 \\
            & & & 3.48$\times$ & 3.25$\times$ & 2.85$\times$ & 2.59$\times$ & 2.39$\times$ \\
            \cmidrule{3-8}
            & & \multirow{2}{*}{\textbf{\sys}} & 588 & 1052 & 1772 & 2736 & 3869 \\
            & & & 4.17$\times$ & 3.80$\times$ & 3.35$\times$ & 2.92$\times$ & 2.55$\times$ \\
            \cmidrule{2-8}
            & \multirow{5}{*}[-0.5ex]{\shortstack[l]{Human-\\Eval}} & AR & 141 & 269 & 532 & 957 & 1684 \\
            \cmidrule{3-8}
            & & \multirow{2}{*}{DSpark} & 445 & 815 & 1464 & 2283 & 3382 \\
            & & & 3.15$\times$ & 3.02$\times$ & 2.75$\times$ & 2.39$\times$ & 2.01$\times$ \\
            \cmidrule{3-8}
            & & \multirow{2}{*}{\textbf{\sys}} & 495 & 914 & 1565 & 2567 & 3654 \\
            & & & 3.50$\times$ & 3.39$\times$ & 2.94$\times$ & 2.68$\times$ & 2.17$\times$ \\
            \midrule
            \multirow{10}{*}[-1.2ex]{Q3-14B}
            & \multirow{5}{*}[-0.5ex]{GSM8K} & AR & 87 & 171 & 330 & 597 & 1020 \\
            \cmidrule{3-8}
            & & \multirow{2}{*}{DSpark} & 355 & 647 & 1142 & 1813 & 2847 \\
            & & & 4.07$\times$ & 3.79$\times$ & 3.47$\times$ & 3.03$\times$ & 2.79$\times$ \\
            \cmidrule{3-8}
            & & \multirow{2}{*}{\textbf{\sys}} & 395 & 721 & 1269 & 1999 & 3010 \\
            & & & 4.53$\times$ & 4.22$\times$ & 3.85$\times$ & 3.35$\times$ & 2.95$\times$ \\
            \cmidrule{2-8}
            & \multirow{5}{*}[-0.5ex]{\shortstack[l]{Human-\\Eval}} & AR & 87 & 170 & 333 & 622 & 1080 \\
            \cmidrule{3-8}
            & & \multirow{2}{*}{DSpark} & 319 & 598 & 1082 & 1688 & 2523 \\
            & & & 3.66$\times$ & 3.52$\times$ & 3.25$\times$ & 2.71$\times$ & 2.34$\times$ \\
            \cmidrule{3-8}
            & & \multirow{2}{*}{\textbf{\sys}} & 335 & 634 & 1140 & 1824 & 2724 \\
            & & & 3.84$\times$ & 3.74$\times$ & 3.42$\times$ & 2.93$\times$ & 2.52$\times$ \\
            \bottomrule
        \end{tabular}
    }
    \vspace{-4mm}
    \label{tab:concurrency}
\end{wraptable}

\paragraph{The advantage persists under batching.}
\sys{} remains faster than DSpark at every measured batch size from
$1$ to $16$ on every evaluated benchmark, for both target models
(Table~\ref{tab:concurrency} shows GSM8K and HumanEval).
The single-request advantage therefore survives batched execution,
rather than disappearing as more requests share each forward pass.
At batch size $16$, \sys{} retains $2.55\times$ and $2.17\times$ speedups
over AR on Qwen3-8B, and $2.95\times$ and $2.52\times$ on Qwen3-14B,
for GSM8K and HumanEval, respectively.
Its throughput exceeds DSpark's by approximately $6\%$--$8\%$ across
these four settings.
Nevertheless, batching reduces the relative benefit of speculation:
absolute throughput increases, but speedup over the correspondingly
batched AR baseline declines for both methods.
Outcome-independent overlap thus does not imply batch-invariant speedup;
the empirical result is a sustained advantage over DSpark and more than
$2\times$ AR throughput throughout the measured range.

\subsection{Further explorations of resource-constrained drafting}
\label{sec:eval:futher}

Dedicating a GPU to drafting lets PSD reduce latency by overlapping
backbone execution with verification, but incurs an additional hardware
cost.
We therefore explore two complementary ways to cut the drafting budget
while retaining \sys{}'s latency advantage: drafting on a much weaker
GPU paired with the H800 target (\emph{\sys{}-A10},
\S\ref{sec:eval:futher:hetero}), and colocating drafting and
verification on one GPU (\emph{\sys{}-single},
\S\ref{sec:eval:futher:single}). Both are compared with the default
two-GPU \sys{}; implementation details are deferred to
Appendix~\ref{app:deployment}.

\begin{figure}[h]
\centering
\includegraphics[width=\textwidth]{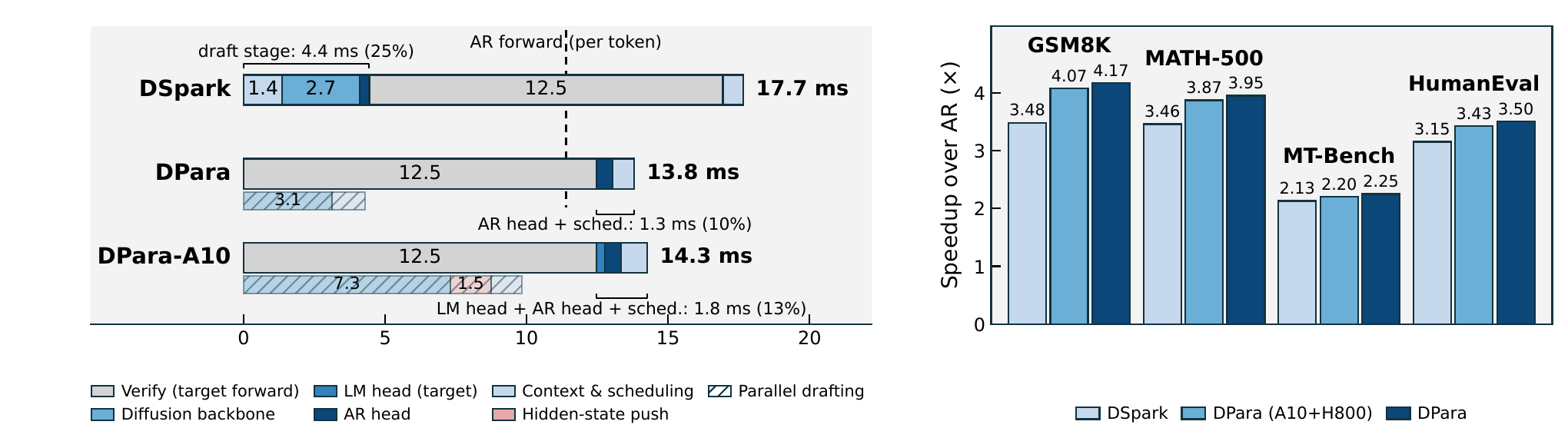}
\vspace{-4mm}
\caption{\textbf{Left}: anatomy of one speculation round on Qwen3-14B
(GSM8K, batch $1$); bars are measured critical paths and hatched
sub-bars show drafting work overlapped with verification. DSpark
serializes its draft stage ($4.4$ ms, $25\%$ of the step);
\sys{} hides the backbone and exposes only $1.3$ ms ($10\%$);
\sys{}-A10 also keeps the A10-side backbone and hidden-state push
inside the verification window, adding only the target-side LM head
($0.3$ ms) to the critical path. \textbf{Right}: Qwen3-8B speedup over
AR at batch size $1$ with the M-DFlash backbone running on an NVIDIA
A10 as the drafting GPU, compared with DSpark and \sys{}.}
\label{fig:main-breakdown}
\end{figure}

\subsubsection{Heterogeneous drafting}
\label{sec:eval:futher:hetero}

\paragraph{Engineering adaptation to the heterogeneous placement.}
\sys{}-A10 pairs the H800 target with an NVIDIA A10 for drafting
(Appendix~\ref{app:deployment}), so the drafting pipeline must be
partitioned across two unequal GPUs. A naive port keeps the whole
pipeline---backbone, LM head, and AR-head sampling---on the A10. Its
AR-head sampling then sits on the critical path, where the latency
negligible on an H800 is magnified on the weaker GPU, motivating a
move to the target. But a target-side AR head must be fed
full-vocabulary logits, an expensive cross-interconnect transfer. We
therefore also move the LM head to the target, so the A10 ships only
the backbone hidden states---roughly $1/30$ the size of the
logits---which cross the link quickly. The residual cost is one
LM-head forward on the target's critical path: we trade a little
compute for a large reduction in network transfer.

\vspace{-4mm}
\paragraph{Impact on end-to-end speed.}
This partition preserves \sys{}'s overlap structure on the weaker GPU
(Figure~\ref{fig:main-breakdown}, left): within a single verification
window the A10 completes both the drafting backbone forward and the
asynchronous transfer of the hidden states of all predicted branches,
leaving only the target-side LM head on the critical path. \sys{}-A10
thus retains about $98\%$ of \sys{}'s speedup on all four benchmarks
and stays well ahead of DSpark (Figure~\ref{fig:main-breakdown},
right); detailed per-step and per-benchmark measurements are in
Appendix~\ref{app:deployment}.

\vspace{-1mm}
\subsubsection{Single-GPU colocation}
\label{sec:eval:futher:single}

\begin{figure*}[t]
\centering
\includegraphics[width=\textwidth]{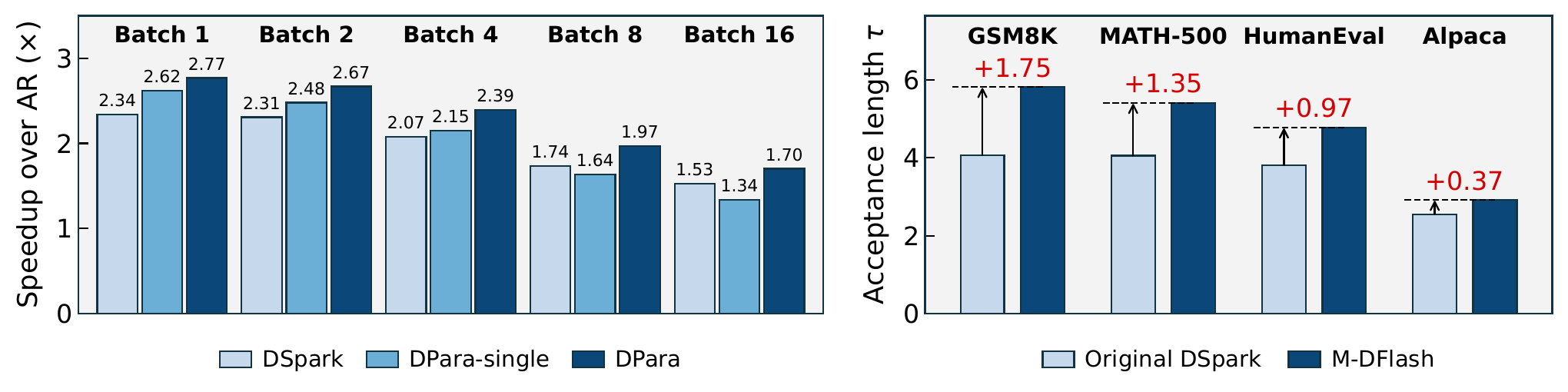}
\vspace{-7mm}
\caption{\textbf{Left}: Qwen3-8B speedup over AR under \sys{}-single and \sys{} with DSpark at different batch sizes; each bar averages the seven benchmarks. \textbf{Right}: acceptance length $\tau$ gains after finetuning on Qwen3-8B's original DSpark checkpoint; both checkpoints run within \sys{} with the same frozen AR head, and red labels show absolute gains.}
\vspace{-2mm}
\label{fig:duo-ablation}
\end{figure*}

\sys{}-single executes verification and backbone precomputation on two
CUDA streams of one GPU, captured in a single CUDA graph
(Appendix~\ref{app:deployment}).
On Qwen3-8B with GB200, it reaches $2.62\times$ speedup over AR at
batch size $1$, compared with $2.34\times$ for DSpark and $2.77\times$
for \sys{}, and retains $90\%$--$95\%$ of \sys{}'s speedup
across batch sizes $1$--$4$ while halving the GPU count
(Figure~\ref{fig:duo-ablation}, left).
The advantage over DSpark disappears at batch sizes $8$ and $16$:
decoding there turns compute-bound, and the colocated backbone
competes with the target for execution resources and bandwidth
(Appendix~\ref{app:deployment}).
Colocation is therefore a resource-efficient option for small batches,
whereas dedicated drafting resources remain preferable as batch size
increases.
\vspace{-1mm}

\setlength{\textfloatsep}{12pt}

\subsection{Ablation Study}
\label{sec:eval:ablation}

\paragraph{Effectiveness of backbone finetuning.}
We test whether finetuning only the backbone adapts the drafter to the
multiple-anchor, feature-less regime required by \sys{}
(\S\ref{sec:design:train}). We run \sys{} on Qwen3-8B with the
original DSpark checkpoint and the finetuned M-DFlash checkpoint,
keeping the AR head and decoding configuration unchanged.
Figure~\ref{fig:duo-ablation} (right) shows that finetuning increases
$\tau$ on all four benchmarks: from $4.07$ to $5.82$ on
GSM8K, $4.06$ to $5.41$ on MATH-500, $3.81$ to $4.78$ on HumanEval,
and $2.55$ to $2.92$ on Alpaca.
These correspond to relative gains of $43.1\%$, $33.3\%$, $25.5\%$,
and $14.4\%$, respectively; the mean $\tau$ across these
four datasets rises from $3.62$ to $4.73$ ($30.7\%$).
The consistency of these gains across math, coding, and chat confirms both
the necessity of adapting the backbone to multiple feature-less anchors and
the reusability of the published AR head without any finetuning.


\section{Conclusion}
\label{sec:conclusion}

We presented \sys{}, the first parallel speculative decoding framework
powered by DSpark-style parallel drafters.
By replacing outcome-dependent
speculation with outcome-independent precomputation, \sys{} eliminates prediction-induced serial
backbone fallback entirely while hiding the dominant drafting cost inside the
verification window. 
This decoupling of backbone execution from
bonus-conditioned token generation reconciles high-capacity drafting with a
minimal critical path, and it translates into state-of-the-art end-to-end
decoding latency across math, coding, and chat workloads.


\bibliography{iclr2027_conference}
\bibliographystyle{iclr2027_conference}

\newpage
\appendix
\section{Latency Breakdown}
\label{app:dflash_time_breakdown}

\paragraph{Profiling setup.}
We profile per-stage latency on the H800 node used for our main results
(\S\ref{sec:eval:setup}), with Qwen3-14B as the target model on GSM8K:
$80$ requests, greedy decoding, draft length $7$, up to $512$ new tokens,
thinking mode disabled. We instrument the verifier call and each drafting component with
per-phase timers and report the mean interval per speculation round over
the $3.6$K--$4.3$K rounds of each run. For \sys, which overlaps drafting
with verification, the overlapped intervals are reported
against the verification window they hide within and are never summed
into the step latency; small residuals between independently timed
intervals are folded into the context and scheduling segments. The same
pattern holds on the other datasets and on Qwen3-8B.

\begin{figure}[h]
\centering
\includegraphics[width=0.55\textwidth]{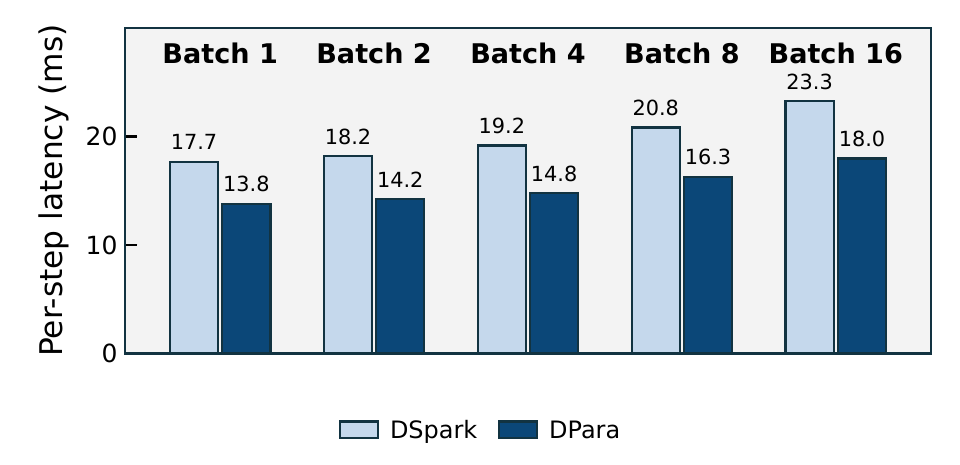}
\caption{Per-step latency across batch sizes on Qwen3-14B (GSM8K, same
profiling setup as Figure~\ref{fig:main-breakdown}). \sys{} is the
lower at every batch size; from batch $1$ to $16$ its step grows
$30\%$, versus $32\%$ for DSpark.}
\label{fig:latency-breakdown}
\end{figure}

\paragraph{Anatomy of one speculation round.}
Figure~\ref{fig:main-breakdown} (left) decomposes a single round at
batch size $1$. Under the conventional draft--verify schedule, DSpark
executes its entire draft stage on the critical path before verification
can start: the diffusion backbone costs $2.7$ ms, the AR head $0.3$ ms,
and context bookkeeping $1.4$ ms, exposing $4.4$ ms---$25\%$ of its
$17.7$ ms step. Across both target models and all seven benchmarks, this
serial draft stage accounts for $24\%$--$32\%$ of the per-step latency
of DFlash/DSpark-style parallel drafters, which is exactly the overhead
that PSD must hide to pay off. DFlash behaves the same way ($4.1$ ms,
$24\%$ of a $17.3$ ms step). \sys{} precomputes the same diffusion
backbone against the verification window and exposes only the
bonus-conditioned AR head plus acceptance bookkeeping: $1.3$ ms ($10\%$)
on top of a $12.5$ ms verify, for a $13.8$ ms step---just $1.21\times$
the AR per-token forward and the lowest per-step latency among the
compared methods. The colocated single-GPU deployment
(\S\ref{sec:eval:futher}) keeps the same structure and still leads
DSpark at $15.1$ ms per step.

\paragraph{Per-step latency across batch sizes.}
Figure~\ref{fig:latency-breakdown} tracks the same quantities
from batch size $1$ to $16$. \sys{} is the fastest at every batch size,
and its step grows only mildly, from $13.8$ to $18.0$ ms ($+30\%$), as
verification lengthens. DSpark follows a similar slope ($17.7
\rightarrow 23.3$ ms, $+32\%$) because its serial draft stage also
lengthens with the batch, but it never closes the gap: its step stays
$28\%$--$30\%$ longer than \sys{}'s at every batch size. Read together
with acceptance length, the per-step gap is decisive: at batch size $1$,
DSpark's higher $\tau$ ($6.5$ vs.\ $5.7$) does not offset its $28\%$
longer step ($2.71$ vs.\ $2.41$ ms per token).

\section{SSD Fallback Rate under Batching}
\label{app:ssd-fallback}

\begin{figure}[h]
\centering
\includegraphics[width=\textwidth]{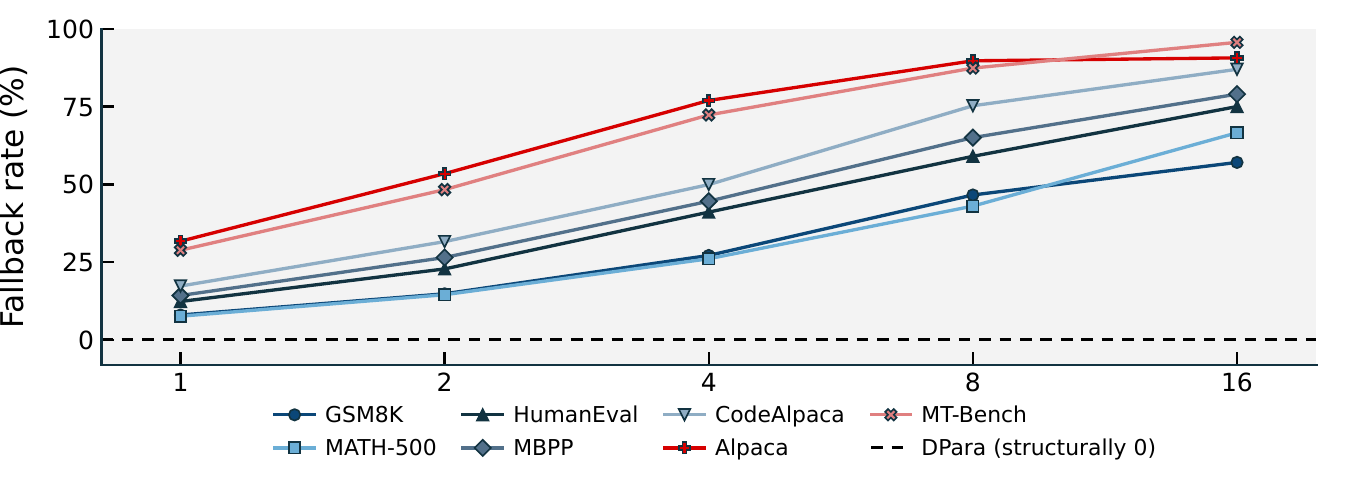}
\caption{Fraction of speculation steps in which SSD falls back to serial
re-drafting, measured on Qwen3-14B (seven benchmarks, batch $1$--$16$,
same configuration as Appendix~\ref{app:dflash_time_breakdown}). A step
falls back when the revealed bonus token of any request in the batch
misses SSD's speculation cache, forcing the drafter to re-draft the
whole batch serially. The rate grows with batch size on every workload
and reaches $57\%$--$96\%$ at batch $16$. \sys{} has no such events at
any batch size.}
\label{fig:ssd-fallback}
\end{figure}

\paragraph{Measuring fallback.}
SSD pre-drafts against a predicted bonus token while the target
verifies (\S\ref{sec:background}). When the revealed bonus token of any
request in the batch misses the speculation cache, the precomputed
continuation is discarded and the drafter re-drafts the whole batch
serially, just in time for the next step. Our profiler instruments
these serial re-drafting calls on the drafting GPU; we report the
fraction of speculation steps that trigger at least one such call,
measured on Qwen3-14B with the configuration of
Appendix~\ref{app:dflash_time_breakdown}.

\paragraph{Fallback rate grows with batch size.}
Figure~\ref{fig:ssd-fallback} shows the measured rates. At batch size
$1$, $8\%$--$32\%$ of steps already fall back, depending on how
predictable the workload's bonus token is; by batch size $16$ the rate
reaches $57\%$--$96\%$. The mechanism is the amplification of a
per-request bonus-mismatch probability $p$ into a per-step fallback
probability $1-(1-p)^B$ for a batch of $B$ requests, since a single
mismatch voids the whole batch's precomputed drafts. Chat workloads
suffer most: their per-request mismatch rate is highest at batch $1$
($29\%$--$32\%$), and by batch $16$ nearly every step is serial
($91\%$--$96\%$), consistent with SSD's weakest speedups on
Alpaca and MT-Bench in Table~\ref{tab:main} and its steep throughput
drop under batching in \S\ref{sec:eval:main}. Math workloads, with the
most predictable bonus tokens ($8\%$ at batch $1$), still fall back in
over half of all steps at batch $16$. \sys{} records no such events at
any batch size: by covering every acceptance boundary instead of
betting on a predicted bonus token, it removes the prediction whose
failure triggers the fallback, so the rate is structurally zero.

\section{Results under Sampling}
\label{app:sampling}

We repeat the batch-size-$1$ comparison of Table~\ref{tab:main} under
sampling: temperature $1.0$ for both the target model and the drafter,
top-$p$ $1.0$, draft length $7$, the same $80$ requests per benchmark,
and up to $512$ new tokens. Acceptance follows the standard
speculative-sampling rejection rule, under which a draft token can be
rejected even when it coincides with the target's most likely
continuation.


\begin{table*}[h]
    \caption{\sys against serial speculative decoding (SD) baselines on Qwen3-8B and Qwen3-14B with sampling decoding. We report decoding speedup over the AR baseline and the average acceptance length $\tau$, with batch size $1$, temperature $1.0$ for both the target and the drafter, up to $512$ new tokens, draft length $7$.}
    \vspace{2mm}
    \label{tab:sampling}
    \resizebox{\linewidth}{!}{
        \normalsize
        \centering
        \setlength{\tabcolsep}{2pt}
        \begin{tabular}{c l @{\hspace{0.6em}} cc cc @{\hspace{0.8em}} cc cc cc @{\hspace{0.8em}} cc cc @{\hspace{0.8em}} cc}
            \toprule
            \multirow{2}{*}{Model} & \multirow{2}{*}{Method} & \multicolumn{4}{c@{\hspace{0.8em}}}{\sc{Math}} & \multicolumn{6}{c@{\hspace{0.8em}}}{\sc{Coding}} & \multicolumn{4}{c@{\hspace{0.8em}}}{\sc{Chat}} & \multicolumn{2}{c}{} \\
            \cmidrule(lr){3-6} \cmidrule(lr){7-12} \cmidrule(lr){13-16}
            & & \multicolumn{2}{c}{GSM8K} & \multicolumn{2}{c@{\hspace{0.8em}}}{MATH-500} & \multicolumn{2}{c}{HumanEval} & \multicolumn{2}{c}{MBPP} & \multicolumn{2}{c@{\hspace{0.8em}}}{CodeAlpaca} & \multicolumn{2}{c}{Alpaca} & \multicolumn{2}{c@{\hspace{0.8em}}}{MT-Bench} & \multicolumn{2}{c}{\sc{Avg.}} \\
            & & \tabsmall Speedup & \tabsmall $\tau$ & \tabsmall Speedup & \tabsmall $\tau$ & \tabsmall Speedup & \tabsmall $\tau$ & \tabsmall Speedup & \tabsmall $\tau$ & \tabsmall Speedup & \tabsmall $\tau$ & \tabsmall Speedup & \tabsmall $\tau$ & \tabsmall Speedup & \tabsmall $\tau$ & \tabsmall Speedup & \tabsmall $\tau$ \\
            \midrule
            \multirow{3}{*}{Q3-8B}
            & DFlash
            & 2.71$\times$ & 5.28 & 2.66$\times$ & 5.15 & 2.39$\times$ & 4.64 & 2.31$\times$ & 4.47 & 2.30$\times$ & 4.46 & 1.56$\times$ & 2.98 & 1.65$\times$ & 3.18 & 2.23$\times$ & 4.31 \\
            & DSpark
            & 3.06$\times$ & 6.15 & 3.00$\times$ & 5.96 & 2.78$\times$ & 5.53 & 2.64$\times$ & 5.24 & 2.63$\times$ & 5.22 & 1.80$\times$ & 3.54 & 1.89$\times$ & 3.74 & 2.54$\times$ & 5.05 \\
            & \textbf{\sys}
            & \textbf{3.45$\times$} & 5.29 & \textbf{3.17$\times$} & 4.77 & \textbf{2.95$\times$} & 4.46 & \textbf{2.83$\times$} & 4.28 & \textbf{2.77$\times$} & 4.20 & \textbf{1.89$\times$} & 2.81 & \textbf{1.97$\times$} & 2.93 & \textbf{2.72$\times$} & 4.11 \\
            \midrule
            \multirow{3}{*}{Q3-14B}
            & DFlash
            & 3.29$\times$ & 5.35 & 3.13$\times$ & 5.04 & 2.86$\times$ & 4.61 & 2.76$\times$ & 4.53 & 2.73$\times$ & 4.45 & 1.85$\times$ & 2.96 & 1.94$\times$ & 3.19 & 2.65$\times$ & 4.30 \\
            & DSpark
            & 3.74$\times$ & 6.18 & 3.61$\times$ & 5.91 & 3.33$\times$ & 5.46 & 3.13$\times$ & 5.25 & 3.15$\times$ & 5.17 & 2.16$\times$ & 3.54 & 2.26$\times$ & 3.79 & 3.05$\times$ & 5.04 \\
            & \textbf{\sys}
            & \textbf{4.00$\times$} & 5.31 & \textbf{3.74$\times$} & 4.88 & \textbf{3.38$\times$} & 4.39 & \textbf{3.14$\times$} & 4.17 & \textbf{3.16$\times$} & 4.15 & \textbf{2.18$\times$} & 2.79 & \textbf{2.28$\times$} & 2.93 & \textbf{3.13$\times$} & 4.09 \\
            \bottomrule
        \end{tabular}
    }
\end{table*}

\paragraph{Sampling lowers acceptance for every method.}
Each method's $\tau$ drops relative to greedy decoding
(Table~\ref{tab:main}): for \sys{}, from $4.43$ to $4.11$ on Qwen3-8B
and from $4.41$ to $4.09$ on Qwen3-14B; DSpark falls from $5.36$ to
$5.05$ and $5.04$, respectively. Average speedups fall
correspondingly: \sys{} goes from $3.21\times$ to $2.72\times$ on
Qwen3-8B and from $3.52\times$ to $3.13\times$ on Qwen3-14B, with
similar relative declines for the serial baselines.

\paragraph{\sys{} retains the best average speedup.}
The ranking of methods is preserved under sampling
(Table~\ref{tab:sampling}). \sys{} keeps the highest average speedup
on both models---$2.72\times$ on Qwen3-8B and $3.13\times$ on
Qwen3-14B, versus $2.54\times$ and $3.05\times$ for DSpark---and is
the fastest on every dataset of both models, including the chat
workload where DSpark tied it under greedy decoding
(Table~\ref{tab:main}). \sys{}'s advantage is orthogonal to the
acceptance rule: its backbone stays hidden inside verification whether
tokens are accepted greedily or by sampling, so the per-step latency
gap over serial drafting
(Appendix~\ref{app:dflash_time_breakdown}) carries over unchanged and
only the acceptance length moves.

\section{Deployment Details for Resource-Constrained Drafting}
\label{app:deployment}

This appendix expands on the two deployments of
\S\ref{sec:eval:futher}.

\paragraph{Heterogeneous role split.}
The naive port of \sys{}'s two-GPU role split places the entire
drafting side, including the M-DFlash LM head and the AR head, on the
drafting GPU. On an A10 this hurts twice: the AR head, which sits on the
critical path, adds substantial latency on the weak GPU, and shipping
full-vocabulary logits crosses the slow interconnect. \sys{}-A10
therefore narrows the drafting GPU to the diffusion backbone alone:
verification, branch selection, the LM head, and AR-head sampling of
the next draft tokens all run on the H800 target. The two machines are
connected over Ethernet (NCCL data plane on TCP sockets, RTT
${\sim}0.8$ ms, measured streaming bandwidth $3.9$--$4.0$ GB/s). The
A10 pushes only the final hidden states of all $d{+}1$ branches across
$d$ positions ($573$ KB for Qwen3-14B, about $1/30$ of the
corresponding logits, ${\approx}0.14$ ms to transfer), and the H800
runs the LM head itself (${\approx}0.3$ ms of extra target-side
compute) before the AR head. Trading cheap target-side compute for
network transfer keeps every network delay off the critical path.

\paragraph{Asynchronous cross-machine transfer.}
Decoding proceeds in verification windows. While the H800 verifies,
the A10 runs the backbone for the next round and asynchronously pushes
the branch hidden states; under full overlap they have all arrived by
the time verification returns, so the H800 can immediately run the LM
head and AR head. It then packs the accepted length and the target
features at the accepted positions into a single asynchronous message
back to the A10, which advances its drafting context and starts the
next backbone pass. Both sides replay captured CUDA graphs to
eliminate scheduling overhead.

\paragraph{Per-step anatomy and end-to-end cost.}
Figure~\ref{fig:main-breakdown} (left) dissects one speculation round
on Qwen3-14B (GSM8K, batch $1$). DSpark serializes its $4.4$ ms draft
stage---$25\%$ of the step---before verification, whereas \sys{} hides
the backbone entirely and exposes only $1.3$ ms ($10\%$). On the A10,
the backbone pass takes $7.3$ ms and the asynchronous hidden-state
push $1.5$ ms; both complete within the $12.5$ ms verification window,
and the target-side LM head adds only $0.3$ ms to the critical path,
for a $14.3$ ms step versus $13.8$ ms for \sys{}. End to end,
\sys{}-A10 retains about $98\%$ of \sys{}'s speedup on all four
benchmarks; averaged over all seven benchmarks, it reaches
$3.14\times$ on Qwen3-8B and $3.43\times$ on Qwen3-14B, versus
$3.21\times$ and $3.52\times$ for \sys{}.

\paragraph{Single-GPU execution.}
We colocate the target and drafter in one process on the same GPU and
execute verification and backbone precomputation on two CUDA streams.
The branch-selection and bonus-conditioned AR-head logic is unchanged.
The head and both forward passes are captured in a single CUDA graph,
and acceptance checks and request scheduling remain outside the graph.
Incremental projection of cached history features avoids repeated work,
while per-request branch state and fixed buffers support safe graph
reuse. Unlike the two-GPU deployment, however, the two streams share
compute and memory bandwidth: logical concurrency does not guarantee
that backbone latency is fully hidden.

\paragraph{Why colocation loses ground at larger batches.}
The single-GPU advantage over DSpark disappears at batch sizes $8$ and
$16$, and its gap to \sys{} widens.
This trend is consistent with a shift from memory-bound toward
compute-bound decoding: larger batches amortize weight reads across
more requests.
At small, memory-bound batches, spare compute capacity allows useful
overlap between drafting and verification. As compute demand grows,
the colocated backbone competes with the target for the same execution
resources and bandwidth, increasing the effective per-step cost;
separate GPUs avoid this direct competition.

\section{Training Recipe}
\label{app:training-recipe}

All M-DFlash backbones are finetuned for one epoch with AdamW under a linear-warmup cosine schedule (learning rate 3e-5, $4\%$ warmup). The effective global batch is $128$ (micro-batch $1$ per GPU with gradient accumulation); each text is truncated to $4096$ tokens and sampled with up to $256$ prediction windows that balance all legal anchor counts $k\in\{1,\dots,d+1\}$. The loss follows \S\ref{sec:design:train}, and target features and supervision are produced online by the corresponding frozen target model. We construct a shared training set from responses generated by Qwen3-8B on prompts from Open-PerfectBlend~\citep{xu2024perfect}, and use it to finetune the M-DFlash drafters for Qwen3-8B and Qwen3-14B. We initialize from the corresponding official DSpark checkpoints and keep the AR heads frozen.

\section{LLM Usage}
We employed LLM-based tools solely during manuscript preparation to refine language and improve readability. The research questions, technical approach, experimental methodology, interpretation of results, and conclusions were developed independently by the authors. The authors reviewed the complete manuscript and remain fully responsible for its content and accuracy.

\end{document}